\documentclass[10pt,twocolumn,letterpaper]{article}

\usepackage[preprint]{cvpr}

\usepackage[pagebackref,breaklinks,colorlinks]{hyperref}

\usepackage[utf8]{inputenc} 
\usepackage[T1]{fontenc}    
\usepackage{url}            
\usepackage{booktabs}       
\usepackage{amsfonts}       
\usepackage{nicefrac}       
\usepackage{microtype}      
\usepackage{xcolor}         
\usepackage{epsfig}
\usepackage{graphicx}
\usepackage{multirow}
\usepackage{dsfont}
\usepackage{algorithm} 
\usepackage{algpseudocode}
\usepackage{url}
\usepackage{breqn}

\usepackage{amsmath,amsfonts,bm}

\def\enc{{q_{\phi}}}
\def\prior{{p_{\theta}}}
\def\dec{{p_{\theta}}}

\def\1{\bm{1}}

\DeclareMathAlphabet{\mathsfit}{\encodingdefault}{\sfdefault}{m}{sl}
\SetMathAlphabet{\mathsfit}{bold}{\encodingdefault}{\sfdefault}{bx}{n}

\newcommand{\lb}{\left[}
\newcommand{\rb}{\right]}

\newcommand{\Loss}{\mathcal{L}}

\newcommand{\ex}[2]{\mathbb{E}_{#1} \lb #2 \rb }

\newcommand{\round}[1]{\ensuremath{\left\lfloor#1\right\rceil}}
\newcommand{\ceil}[1]{\ensuremath{\left\lceil#1\right\rceil}}

\def\cvprPaperID{*****}
\def\confName{CVPR}
\def\confYear{2022}

\title{Implicit Neural Video Compression}

\begin{document}


\author{%
    Yunfan Zhang
    \quad
    Ties van Rozendaal
    \quad
    Johann Brehmer
    \quad
    Markus Nagel
    \quad
    Taco S.~Cohen \\
    Qualcomm AI Research\thanks{Qualcomm AI Research is an initiative of Qualcomm Technologies, Inc.}\\
    {\tt\small \{yunfzhan, ties, jbrehmer, markusn, tacos\}@qti.qualcomm.com}
}

\maketitle

\begin{abstract}
 We propose a method to compress full-resolution video sequences with implicit neural representations. Each frame is represented as a neural network that maps coordinate positions to pixel values. We use a separate implicit network to modulate the coordinate inputs, which enables efficient motion compensation between frames. Together with a small residual network, this allows us to efficiently compress P-frames relative to the previous frame. We further lower the bitrate by storing the network weights with learned integer quantization. Our method, which we call \emph{implicit pixel flow}~(IPF), offers several simplifications over established neural video codecs: it does not require the receiver to have access to a pretrained neural network, does not use expensive interpolation-based warping operations, and does not require a separate training dataset. We demonstrate the feasibility of neural implicit compression on image and video data. 
\end{abstract}

\section{Introduction}
\label{intro}

Video streaming makes up a major portion of today's internet traffic~\cite{internet_traffic}. Compression codecs based on deep learning~\cite{lu2019dvc, agustsson2020scale} have recently become competitive with popular classical codecs like H.264 (AVC)~\cite{x264} and H.265 (HEVC)~\cite{x265}, but these methods have not yet been widely adapted in real-life applications. One reason for this is that neural codecs do not yet robustly outperform traditional codecs in terms of compression performance. At least as important, however, are practical considerations: neural codecs require access to a (typically large) neural network on each device on which videos need to be decompressed. This is memory-heavy, difficult to maintain, and can be vulnerable to corruption. A lightweight, computationally efficient neural codec that does not require storing large network weights might be more practical, especially for on-device applications. Moreover, standard neural codecs require a training dataset that is similar to the video samples expected at test time; the compression performance potentially suffers under training set bias and domain shift, for instance when networks trained on natural scene data are used to compress animated sequences.

We propose \emph{implicit pixel flow} (IPF), a new method for video and image compression based on implicit neural representations (INR) that addresses these practical shortcomings. Each frame is represented as a function that maps coordinates within the frame to RGB values. We implement these functions as neural networks, building on recent progress in neural scene representation. Encoding then consists of choosing the architecture and overfitting the network weights on the video frames. Decoding only requires forward passes of the network. We quantize the neural network weights with fixed-point integer quantization with learned parameters and separate per-channel bit widths.

To further reduce the bitrate for video data, we compress most frames as P-frames, i.\,e.\ using the information from the previous frame. We leverage the similarity of subsequent frames through a separate implicit network that outputs the optical flow field, a map of displacement vectors that can be applied to the previous frame to approximate the current frame. We argue that implicit neural representations are a natural fit for such an optical flow warping operation: they require a simple addition in the input space, avoiding the usual interpolation-based operations that are computationally expensive and difficult to implement on device~\cite{lu2019dvc, agustsson2020scale}. In addition to the lightweight flow network, we train an equally lightweight residual network to complete the modeling of a P-frame.

One key advantage of our codec is that it eliminates the need to store a neural network on the receiver side, instead it only requires a framework for the evaluation of neural networks to be present. Due to the efficient flow warping operation, decoding is less computationally expensive and more hardware-friendly than established neural video codecs. Finally, this method does not require any separate training dataset. Not only does this avoid potential privacy concerns, it also means that our codec performs well on data from different domains, including those for which no suitable training data is available.

\section{Related Work}
\label{related_work}

\paragraph{Neural compression codecs}
Neural video compression typically follows the framework of variational~\cite{kingma2013auto} or compressive~\cite{Theis2017-qu} autoencoders consisting of an encoder $q$ that maps images or video frames to a latent variable $z$, a decoder $p$ that maps the latent variables to the reconstructed image, and a prior $p(z)$ under which the quantized latent variables are entropy-coded. On a training dataset, these models learn to minimize the rate-distortion ($RD$) loss
\begin{dmath}
    \Loss_{RD}(\phi, \theta)
    =
    \mathbb{E}_{x\sim p(x)} \biggl[
    \beta \underbrace{\ex{z\sim\enc(z|x)}{- \log \prior(z)}}_{R}
    + \underbrace{\ex{z\sim\enc(z|x)}{-\log \dec(x | z)}}_{ D}
    \biggr] \,.
    \label{eq:RD_loss}
\end{dmath}
The trained prior and decoder need to be available at the receiver side in order to decode the transmitted bitstream.  Works following this scheme include Refs.~\cite{Habibian2019-oe, pessoa2020end}, which use 3D convolution architectures, and Refs.~\cite{agustsson2020scale, Chen2018-wj, liu2020learned, NIPS2019_9127, lu2019dvc, rippel2019learned, golinski2020feedback, Cheng_2019_CVPR, choi2019deep, Djelouah2019-ij, Wu2018-bu, park2019deep, pourreza2021extending, rippel2021elf, hu2021fvc}, which model P-frames as an optical flow field applied to the previous frame plus a residual model.

Recently, Ref.~\cite{van2021overfitting} introduced instance-adaptive fine-tuning, in which the full autoencoder model is fine-tuned on each test instance. The network weight updates are entropy-coded to the bitstream and transmitted alongside the latents. While this approach relaxes the requirements on the model to generalize from the training dataset to any instance encountered at test time, it still requires a pretrained decoder and prior model to be available at the receiver side.

The first publication to apply implicit neural representation to compression is Ref.~\cite{dupont2021coin}. The authors propose to compress images through their implicit representation as neural network weights. They focus on SIREN models~\cite{sitzmann2020siren} with varying numbers of layers and channels and quantize them to 16-bit precision. 

In this work, we propose a video codec that features an I-frame codec with several improvements over Ref.~\cite{dupont2021coin}. Namely, we share the compute across neighboring pixels to reduce complexity and deploy a learned quantization scheme that can achieve better bitrate efficiency, and that can be trained end-to-end using a rate-distortion loss. 

During the final preparations of this manuscript, two concurrent works appeared. In Ref.~\cite{chen2021nerv} the authors train a single model over the entire video. For the spatial dimensions, the authors use an upsampling procedure similar to ours, achieving faster training and inference speeds. The authors then use pruning as well as quantization to reduce the bit-rate of the trained networks. The resulting RD performance is impressive, matching traditional codecs at low bitrates. However, such models require large chunks of a video to be encoded at once and cannot be applied in a low-delay setting. On the other hand, Ref.~\cite{struempler2021implicit} introduces an image compression codec based on quantization-aware training and meta-learned initializations. The result shows improvement over JPEG as well as our base models.

\paragraph{Implicit neural representations}
Implicit representations have been successfully used for learning three-dimensional structures~\cite{mescheder2019occupancy,chen2019learning,deng2020nasa,park2019deepsdf,atzmon2020sal,genova2019learning,genova2020local,jiang2020local} and light fields~(see \cite{yariv2020multiview,mildenhall2020nerf,niemeyer2020differentiable,park2021nerfies,liu2019learning,liu2020dist,li2021neural,sitzmann2019scene} and references therein).
Similar to our approach to compression, these works train a neural network on a single scene such that it is encoded by the network weights. New views of the scene can then be generated through a forward pass of the network. These methods are proven to be more efficient than their discrete counterparts due to their ability to exploit the redundancies.

While implicit representations have also been applied to data with lower-dimensional coordinates such as images and videos~\cite{stanley2007compositional,sitzmann2020siren,tancik2020fourier,mehta2021modulated}, the relative efficiency compared to discrete or latent representations has not been carefully studied. Furthermore, implicit representations have to compete with established compression codecs. Reference~\cite{dupont2021coin}, to the best of our knowledge the only comparison of implicit image compression methods to classical codecs yet, demonstrated that this is no easy task.

Regardless of the dimension of input data, choosing the right class of representations is important. It has been shown that Fourier-domain features are conducive to implicit neural models learning the structure of realistic scenes. Specifically, Ref.~\cite{tancik2020fourier} proposed using randomly sampled Fourier frequencies as encoder prior to passing in MLP model. Ref.~\cite{sitzmann2020siren} showed that pixel-space MLPs can achieve comparable results when using sinusoidal activations, provided that the weights are initialized carefully.

\paragraph{Dynamic scene representations}
Implicit representations are continuous in nature. Shifting the input to these networks corresponds to continuous spatial translations within the represented scene or image. In this way, implicit representations lend themselves to tasks such as super-resolution or warping of 3D scenes~\cite{park2021nerfies,pumarola2021dnerf,du2021nerflow,xian2021spacetime,li2021neural}. We extend this idea to motion compensation between video frames and propose to compress the displacement map with another implicit neural representation. The closest related work we are aware of is Ref.~\cite{park2021nerfies}, which introduces an auxiliary network to model dynamic warping between smartphone selfies. They introduce a reference coordinate frame for each scene and compute the warping of each picture with respect to the reference. Unlike selfies, videos frames are inherently sequential. We take advantage of this natural ordering in designing our method.

\paragraph{Model quantization}
Neural network quantization is an active field of research with the goal of reducing the size of models and running them more efficiently on resource-constrained devices. Good surveys of the field are~\cite{nagel2021whitepaper, gholami2021survey}.
On a high level, there are two lines of work: \emph{vector quantization}~\cite{han2015deep_compression, Stock2020And}, which represents a quantized tensor using a code book, and \emph{fixed-point quantization}~\cite{krishnamoorthi, dfq, nagel2020up, lsq}, which represents a tensor with a fixed-point number that consists of a integer tensor and a scaling factor.
In this paper we focus on the latter as it enables memory savings and leads to reduced compute complexity during training. The fixed-point quantization function is defined as
\begin{equation}
    \label{eq:quantization}
    Q_{s,b}(\theta) := s \cdot \theta_{\text{int}} = s \cdot \mathrm{clamp}\left(\round{\frac{\theta}{s}}; 2^{b-1},2^{b-1} - 1 \right) 
\end{equation}
where $\theta_{\text{int}}$ is a integer tensor with $b$ bits and $s$ is a scaling factor (or vector) in floating point. We will use the symbol $\tau = (s, b)$ to refer to the set of all quantization parameters.

Low-bit quantization of all weight tensors in a neural network can incur significant quantization noise.
With quantization-aware training~\cite{jacob2018cvpr, pact2018, lsq, bhalgat2020lsq+}, neural networks can adapt to the quantization noise by training them end-to-end with the quantization operation.
As the rounding operation in Eq.~\eqref{eq:quantization} is non-differentiable, commonly the straight-trough estimator (STE)~\cite{bengio2013estimating} is used to approximate its gradient.
Next to learning the scaling factors jointly with the network, recent work also started on learning a per-tensor bit-width for every layer~\cite{differentiablequantization, bayesianbits}.
We extend the approach of \cite{differentiablequantization} to learn a per-channel bit-width and scaling factor.
Contrary to earlier approaches, we formulate the quantization bit-width as a rate loss, and minimize the $RD$-loss (Eq.~\eqref{eq:RD_loss}) to learn the best trade-off between bitrate and distortion in pixel space.

\section{Implicit pixel flow}
\label{methods}

\subsection{Overview}

Our proposed video compression codec is based on representing images through neural networks that map coordinates to RGB values. The compressed file thus consists of a small header that specifies the network architecture, and the weights of a neural network for each frame. On the one hand, an expressive architecture ensures that the network can learn the image with high fidelity. On the other hand, we quantize the network weights to keep its size down.

To compress video data more efficiently, we make use of the similarity of successive frames. We split the video in small blocks of frames (``groups of pictures'' or GoP). The first frame in a GoP is compressed as an I-frame, training a single network to compress an image. The remaining frames in the GoP are trained as P-frames, i.\,e.\ using the previous frame as reference.

For P-frames, we introduce two ideas that reduce the bitrate. First, we introduce a separate implicit representation of the optical flow field or estimated motion vector, which describes how much different parts of the scene moved between frames. By adding this flow field to the inputs of our neural representation, we get a sensible first approximation of the current frame. This form of motion compensation (or optical-flow warping) is a much simpler and less computationally expensive operation than the interpolation-based approach typically used in neural video codecs~\cite{agustsson2020scale}. Second, we introduce a residual model that further improves the distortion. 

Encoding a video consists of training a network with quantized weights, minimizing the rate-distortion loss
\begin{multline}
    \Loss_{\textup{IPF}}(\theta, \tau, \omega)
    =
    \underbrace{\mathbb{E}_{t,x,y} \lVert f_{Q_{\tau_t}(\theta_t)}(x,y) - I_{t, x,y} \rVert_2^2}_{D} \\
    + \beta \underbrace{\mathbb{E}_{t} \sum_i b_{i,t}}_{R} \,.
    \label{eq:implicit_RD_loss}
\end{multline}
Here $t$ is the frame index, $x, y$ are the coordinates within a video frame and $I_{t, x,y}$ are the ground-truth RGB values at these coordinates. $f_{\theta_t} (x,y)$ is the implicit neural network with weights $\theta_t$ evaluated at coordinates $(x,y)$. $Q_{\tau_t}$ is the quantization function with parameters $\tau_t$,
and $b_{i,t}$ are the learned bit sizes of the parameters, a function of quantization parameters $s_i$ and $\theta_{\mathrm{max}\,i}$ (to be defined in Sec. \ref{sec:method_quantizer_and_weight_prior}). The rate term $R$ corresponds to the rate term in Eq.~\eqref{eq:RD_loss} with a uniform prior over quantized weights.
Decoding a video file initially requires reading the architecture header, reading the neural network weights frame by frame, and evaluating the networks on the required pixel coordinates (which may be on a regular grid, but can also deviate from that).

In the following, we will discuss in more detail our image representation in Sec.~\ref{sec:implicit_image_representations}, followed by the video structure and the optical flow warping in Sec.~\ref{sec:implicit_video_representations}. In Sec.~\ref{sec:training} we discuss the training pipeline in more detail, before finally describing the quantization and entropy coding scheme in Sec.~\ref{sec:method_quantizer_and_weight_prior}.

\subsection{Implicit image representations}
\label{sec:implicit_image_representations}

\paragraph{SIREN}
At the heart of our compression codec is the choice of neural network architecture used to represent individual images. Like most neural implicit models, we use models that take coordinates within an image as input and return RGB values,
\begin{equation}
    f : \mathbb{R}^2 \to \mathbb{R}^3 \,, \quad
    (x, y) \to f_\theta(x,y) =
    \begin{pmatrix}
        r(x,y) \\
        g(x,y) \\
        b(x,y)
    \end{pmatrix} \,.
\end{equation}
The most common models used in the literature are multi-layer perceptrons (MLP). Specifically, we find that the SIREN architecture allow the highest expressivity~\cite{sitzmann2020siren}. By using periodic activation functions, these models ensure that fine details in images and videos can be represented accurately. Decoding an image entails evaluating the MLP at every pixel location $(x,y)$ of interest.

The input space of this representation is continuous. We are thus not tied to any particular pixel grid, and the representation can be trained or evaluated at different resolution settings or on irregular grids. Our method is therefore suitable for super-resolution tasks or for decoding video data directly in a target resolution depending for instance on the available hardware.

\paragraph{Upsampling SIREN (uSIREN)}
While a SIREN-based MLP~\cite{sitzmann2020siren} is expressive, it requires one forward pass for each pixel in decoding an image, which can quickly get expensive on full resolution media. To lighten the decoding compute, we share part of the computation between neighboring pixels. To see how this works, notice that a MLP can also be regarded as a $1 \times 1$ convolution, where the convolving dimensions are the coordinate dimensions of the input. By inserting a bilinear interpolation layer of stride $n$ between the $1 \times 1$ convolution layers, we reduce the operating dimension or layers preceding the upsampling by a factor of $n^2$. This means the decoding compute of those layers are also reduced by a factor of $n^2$.

\subsection{Implicit video representations}
\label{sec:implicit_video_representations}

\paragraph{Overview}
Video data often have strong redundancies between subsequent frames. Neural implicit representations can represent video data by extending the input space by a third time or frame dimension~\cite{xian2021spacetime,mehta2021modulated}. While this approach is straightforward, we find that the implicit networks are not expressive enough to represent high-resolution video data at low distortion, as we demonstrate in experiments in the supplementary materials. In addition, this 3D approach in less suitable for video streaming, because an entire block of frames needs to be received before even the first frame can be decoded.

Instead, we propose to compress video sequences frame by frame while still leveraging the similarity between them.
Like many classical and neural video codecs, we split the full video sequence into multiple GoPs.
In each GoP, only the very first frame is compressed as a stand-alone I-frame, while all other frames are compressed using available information from other frames. Here we focus on the common case where each frame in a GoP except the first is compressed as a P-frame: it explicitly depends on the previous frame, but not future frames. This setup is well-suited to video streaming in a low delay setting.

\begin{figure}[t]
    \centering
    \includegraphics[width=0.45\textwidth]{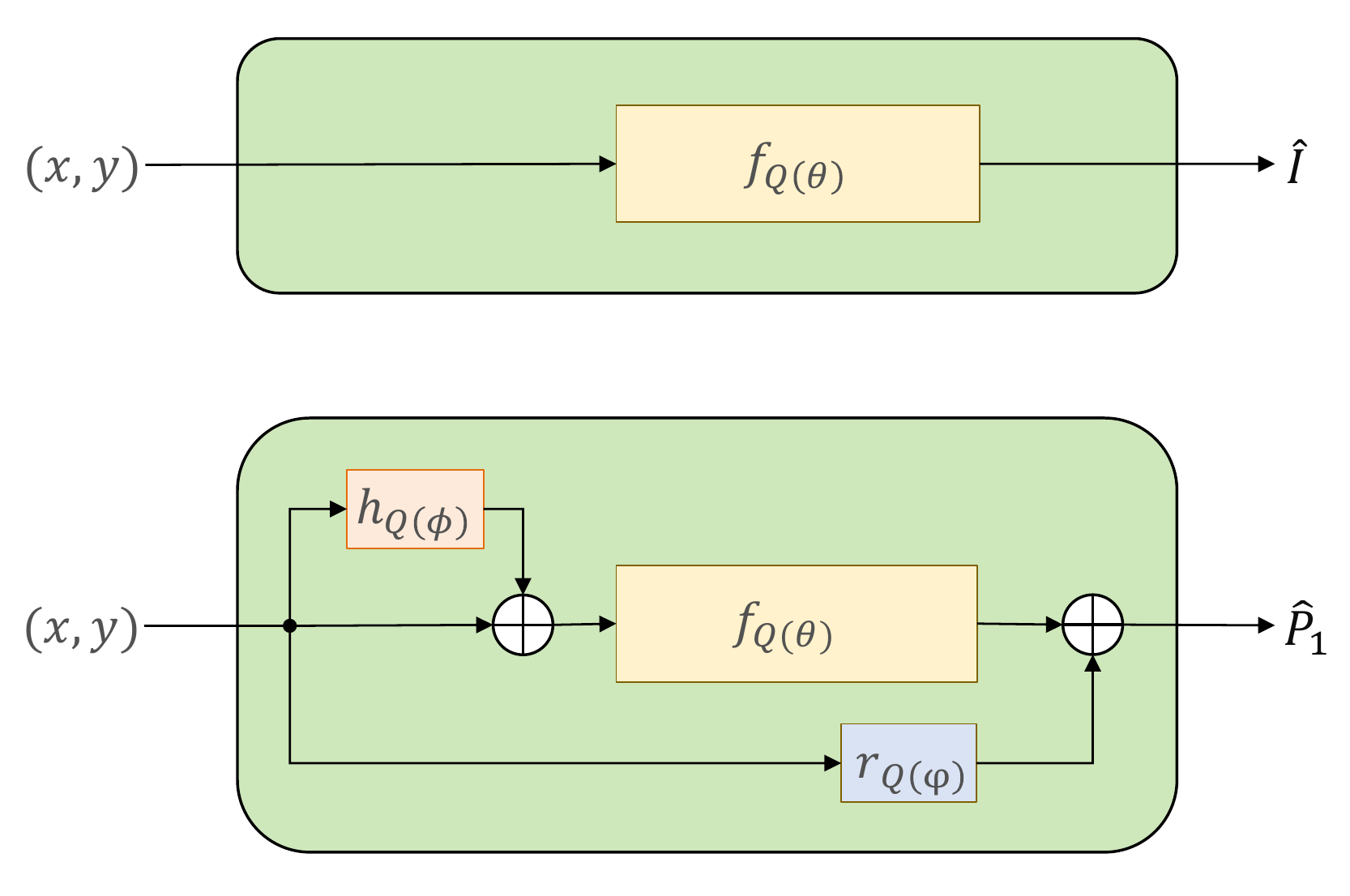}%
    \caption{Details of the operations inside our codec. The top block shows an I-frame. The bottom block depicts a P-frame, where we take the I-frame implicit model $f$, add a "flow"-field modeled by a network $h$ to the inputs and optionally add a residual modeled by a network $r$ to the output.}
    \label{fig:pipeline}
\end{figure}

One popular strategy to compress P-frames is to estimate the optical flow field or motion vector, a field of displacement vectors that, applied to the previous frame, give a good first estimate of the current frame. In addition to the optical flow, pixel-level~\cite{lu2019dvc, agustsson2020scale} or feature-level~\cite{hu2021fvc} residuals are compressed. The final prediction for the frame is given by applying the optical flow field to the previous reconstructed frame in an operation known as warping or motion compensation and adding the residuals. While this approach can substantially lower the bitrate, optical-flow warping requires expensive and hardware-unfriendly interpolation operations.

\begin{algorithm*}
	\caption{Training procedure}
	\label{alg:1}
	\begin{algorithmic}[1]
        \State Write hyperparameters to bitstream
		\For {$\mathrm{GoP} = 1, 2, \dots$}
            \State Pre-train full-precision base model $f_{\theta_0}$ on distortion term of objective \eqref{eq:implicit_RD_loss} on I-frame \Comment{I-frame training}
            \State Train quantized model $f_{Q_{\tau}(\theta_0)}$  with rate-distortion objective \eqref{eq:implicit_RD_loss} on I-frame
            \State Write learnable quantization parameters $\tau$, and quantized I-frame weights $\theta_0$ to bitstream
    		\State Initialize displacement $\Delta_0 \leftarrow 0$ \Comment{P-frame training}
    		\For {$t=1,2,\ldots$}
    			\State Load existing model with parameters $\theta,\Delta_{t-1}$
    			\State With $\theta$ frozen, pre-train $\psi_t$ and $\phi_t$ with distortion objective \eqref{eq:complete_p_frame_model} on frame $P_t$
    			\State Train quantized models on rate-distortion objective and write updates $\psi_t$ and $\phi_t$ to bitstream.
    			\State Update displacement tensor $\Delta_t \leftarrow \Delta_{t-1}+h_{\phi_t}$
    		\EndFor
		\EndFor
	\end{algorithmic} 
\end{algorithm*}

\paragraph{Implicit flow warping}
In this work we model optical flow implicitly by leveraging the fact that implicit representations are continuous. Recall that frames are represented as a network that takes image coordinates as input, $(x,y) \to f_t(x, y) = f_{\theta_t}(x,y) = (r,g,b)$. Applying the displacement from an optical flow field $h_\phi(x,y) = (\Delta_t^x, \Delta_t^y)$ requires only to add the displacement vector to the input variables:
\begin{align}
    (x, y) \to f_t(x,y) &= f_{t-1}  \circ (1 + h_{\phi_t}) (x, y) \notag \\
    &= f_{t-1}(x + \Delta_t^x, y + \Delta_t^y) \,.
    \label{eq:warping}
\end{align}
The displacement fields $(\Delta_t^x, \Delta_t^y)(x, y)$ are represented implicitly as neural networks $h_{\phi_t}$ with weights $\phi_t$, using smaller SIREN architectures (Sec.~\ref{sec:implicit_image_representations}).

\paragraph{Residual modeling}

Even an accurate flow model cannot always fully predict a frame, for instance due to occlusion effects or the introduction of new objects. We therefore model residuals on top of the warped frame, 
\begin{equation}
    (x, y) \to f_t(x,y) = f_{t-1} \circ (1 + h_{\phi_t}) (x, y) + r_{\psi_t}(x, y) \,,
    \label{eq:warping_residual}
\end{equation}
with a separate implicit network $r_{\psi_t}(x, y)$. The same small model as for the flow suffices for a good performance, allowing us to keep the bitrate low. In ablation studies we found this modeling of residuals in pixel space to be more effective than modeling residuals in the weight space of the implicit networks.

\paragraph{Sometimes, implicit flow is all you need}

Like many autoencoder-based video codecs, our IPF model thus follows the basic recipe of modelling P-frames as the previous frame warped by a compressed flow field plus a compressed residual. However, the learning dynamics lead to some practical differences: unlike in the autoencoder approach, the bitrate spent on flow fields is largely determined by the model architecture, which is fixed prior to training. Therefore there is less of an incentive for the flow field to be smooth and to consist of large near-constant patches. Instead, the IPF flow can model changes in fine-grained detail compared to the previous frame. In our experiments, this works so well that for many frame transitions the residual model is redundant.\footnote{We find this behaviour independent of whether we train flow and residual model jointly or sequentially.} Of course, this does not work when pixels in new colors appear, for instance because of new objects entering the scene.

We embrace this expressiveness of the flow and explore a optimized codec where the residual model is optional. In this case we dynamically decide on the inclusion of a residual model on a per GoP basis. On the encoder side, for each GoP we compute the rate and distortion performance both with and without the residual model. We choose the variant that leads to a lower $RD$ loss and signal this choice by transmitting a single bitstream. If flow without residuals is the better choice for a frame, we do not need to transmit the residual model, reducing the bitrate.

\subsection{Quantization and entropy coding}
\label{sec:method_quantizer_and_weight_prior}

To reduce the model size of the implicit models representing I-frames, optical flow, and residuals, we quantize every weight tensor $\theta^{(l)} \in \theta$ using fixed-point representation (cf.\ Eq.~\eqref{eq:quantization}). To learn the quantization parameters and bit-width jointly with the model weights, we follow the parameterization suggested by Ref.~\cite{differentiablequantization} and learn the scale $s$ and the clipping threshold $\theta_{\text{max}}$. The bit-width $b$ is then implicitly defined as 
\begin{equation}
    b(s, \theta_{\text{max}}) = \log_2 \left(\ceil{\frac{\theta_{\text{max}}}{s}} + 1 \right) + 1 \,.
\end{equation}
Reference~\cite{differentiablequantization} showed that this parameterization is favorable over learning the bit-width directly as it does not suffer from an unbounded gradient norm.
We further extend this approach to per-channel quantization~\cite{krishnamoorthi} allowing us to learn a separate range and bit-width for every row in the matrix\footnote{Output channel in case of a convolutional layer.}. Our per-channel mixed precision quantization function is defined as:
\begin{equation}
    Q_\tau(\theta_{ij}) = 
    \begin{cases}
        s_{i} \cdot \round{\frac{\theta_{ij}}{s_i}}         & \quad |\theta_{ij}| \leq \theta_{\text{max},i},\\
        \text{sign}(\theta_{ij}) \cdot \theta_{\text{max},i}   & \quad |\theta_{ij}| > \theta_{\text{max},i}.
    \end{cases}
\end{equation}

Next, we encode all quantization parameters $\tau = \{s^{(l)}, b^{(l)}\}_{l=1}^L$  and all integer tensors $\theta_{\text{int}}=\{\theta^{(l)}_{\text{int}}\}_{l=1}^L$ to the bitstream. The $s^{(l)}$ are encoded as 32-bit floating point vectors, the bit-widths $b^{(l)}$ as 5-bit integer vectors, and the 
$\theta^{(l)}_{\text{int}}$ in their respective per-channel bit-width $b^{(l)}_i$.

\begin{figure*}
    \centering%
    \includegraphics[width=0.85\textwidth]{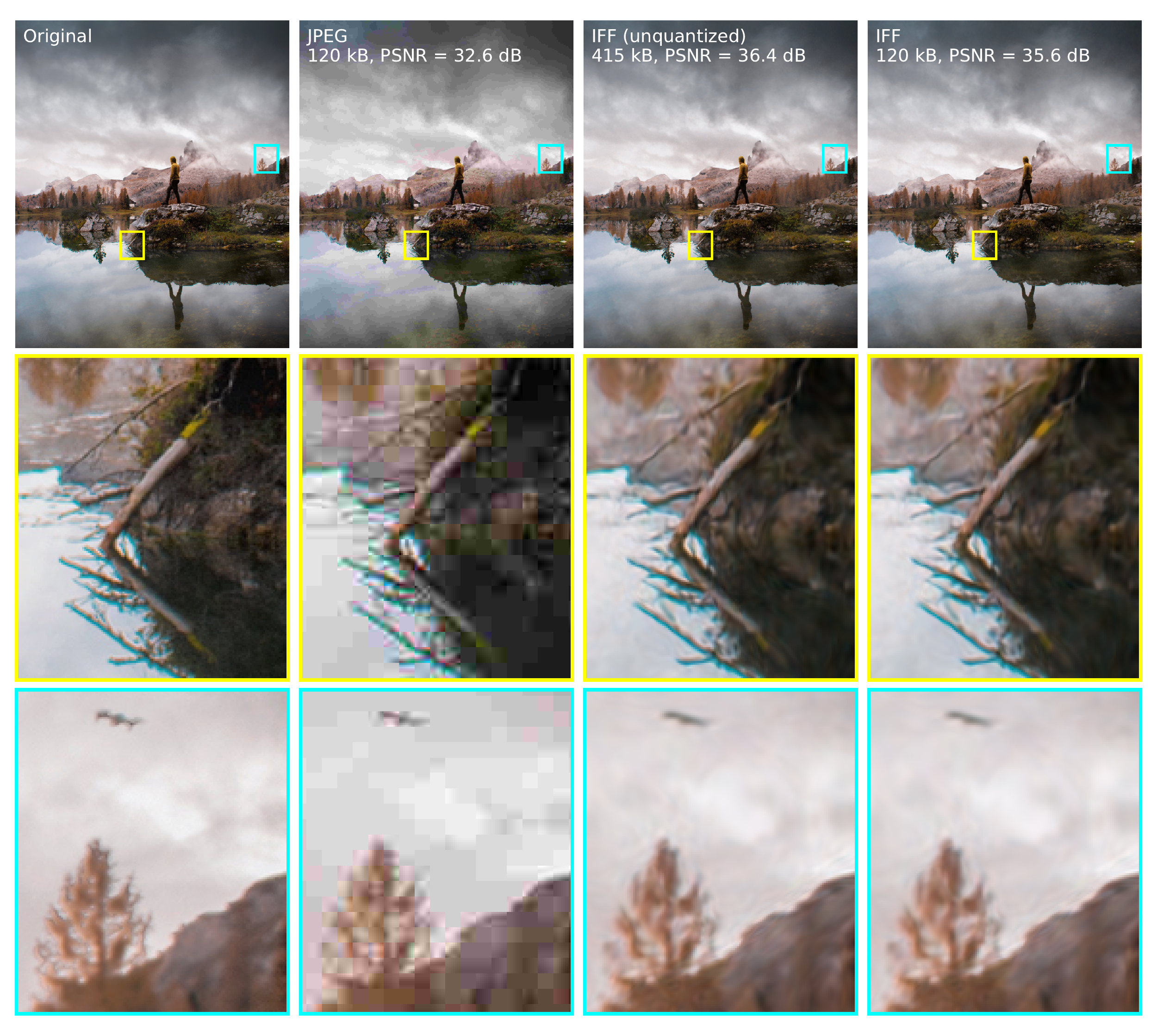}
    \caption{Uncompressed image from the CLIC 2020 Professional dataset~\cite{clic} (left) compared to images compressed with JPEG and our neural implicit video codec (IPF). For comparison, we also include an unquantized implicit representation, in which all weights are stored at 32-bit floating-point precision. The top row shows the full image, the other rows zoom in on details.}
    \label{fig:perceptual}
\end{figure*}

\subsection{Training pipeline}
\label{sec:training}

We summarize our architecture in Fig.~\ref{fig:pipeline} and detail the training procedure in Alg.~\ref{alg:1}. Initially, unquantized networks are pre-trained by minimizing distortion, then the quantizers are added and the quantized models are trained on the combined rate-distortion loss of Eq.~\eqref{eq:implicit_RD_loss}. The quantized weights are written to the bitstream.

We train the base model for each I-frame, and flow and residual models for each P-frame. While each residual model is independent, we find it beneficial to accumulate the flow across frames inside a GOP. This way each flow model only needs to model the motion between consecutive frames, instead of directly from the I-frame.

Unrolling the recursive definition of warping, the implicit representation of the P-frame $t$ is given by
\begin{align}
    f_{t}(x,y)
    &=  f_{0} \circ (1 + h_{\phi_t}) \circ \dots \circ (1 + h_{\phi_1}) (x, y)
    + r_{\psi_t}(x,y) \notag \\
    &= f_{0}\Bigl(x + \sum_{i=1}^{t-1} \Delta_i^x + h^x_{\psi_t}(x,y), \notag \\
    &\phantom{=}\quad \quad  y + \sum_{i=1}^{t-1} \Delta_i^y + h^y_{\psi_t}(x,y)\Bigr)
    + r_{\psi_t}(x, y) \,, \!\!
    \label{eq:complete_p_frame_model}
\end{align}
where for readability we leave out the quantizer and $f_0(x,y)$ is the implicit parameterization of the previous I-frame. In the second line, we show that we do not need to store the previous flow networks $h_i$ in memory and evaluate them again for each frame; instead, we just need to store the cumulative flow field $\sum \Delta$ in a single tensor. This tensor can be constructed in the same way on receiver side. Training the flow and residual networks for a P-frame then just amounts to minimizing the loss in Eq.~\eqref{eq:implicit_RD_loss} with the parameterization in Eq.~\eqref{eq:complete_p_frame_model}.

\section{Experiments}

We now demonstrate our compression codec in experiments. First we demonstrate the ability of our neural implicit model to represent images, before turning to video data.

\paragraph{Image compression}
In Fig.~\ref{fig:perceptual} we show an image from the CLIC 2020 challenge, a version compressed with JPEG, and a version compressed with our neural implicit base model (see the supplementary materials for a detailed description of the setup).  Both codecs are run at a strong compression setting, the image with resolution $1716 \times 2048$ is in both cases compressed to a 200\,kB file. We confirm that the network can represent the image including fine-grained detail. Compared to JPEG, it achieves a substantially better PSNR at the same file size. At this low-filesize setting, there are certainly artifacts, but they are less pronounced than those induced by JPEG.

\begin{figure}[t]
    \centering
    \includegraphics[width=0.42\textwidth]{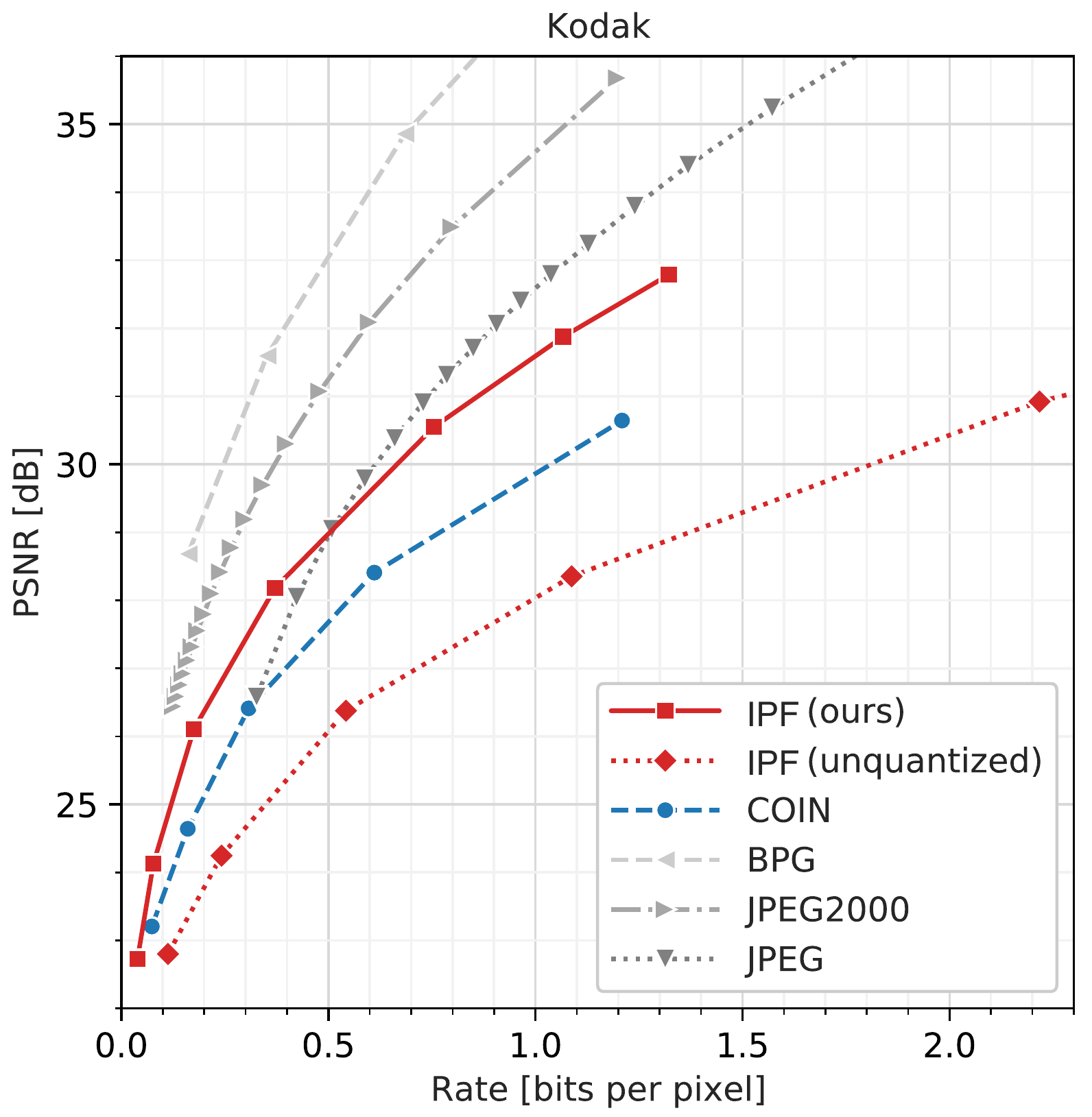}%
    \caption{Image compression performance of our IPF codec (red solid) compared to COIN, JPEG, JPEG2000, and BPG baselines on the Kodak test image dataset~\cite{kodak}. The dashed line indicates the performance without learned integer quantization, storing all network weights at 32-bit floating point precision.}
    \label{fig:kodak_rd}
\end{figure}

In addition, we compare to an unquantized implicit representation with network weights stored at 32-bit floating-point precision. Quantization is able to reduce the network size substantially at a small cost in the distortion performance. Note that we are able to quantize the parameters to an average of 9.3 bits per parameter without a substantial drop in distortion performance.

In Fig.~\ref{fig:kodak_rd} we evaluate the image compression performance on the Kodak image dataset~\cite{kodak}. The rate-distortion results in Fig.~\ref{fig:kodak_rd} show that our method outperforms JPEG at low to medium bitrates, but is not competitive with state-of-the art codecs like BPG. We also compare to COIN~\cite{dupont2021coin}, which uses very similar implicit networks, but a simpler weight compression scheme. We find that our learned integer quantization, which allows us to compress the networks to around 9 to 10 bits per parameter, leads to a substantial performance improvement over COIN with its 16-bit floating point quantization.

\begin{figure}[t]
    \centering
    \includegraphics[width=0.45\textwidth]{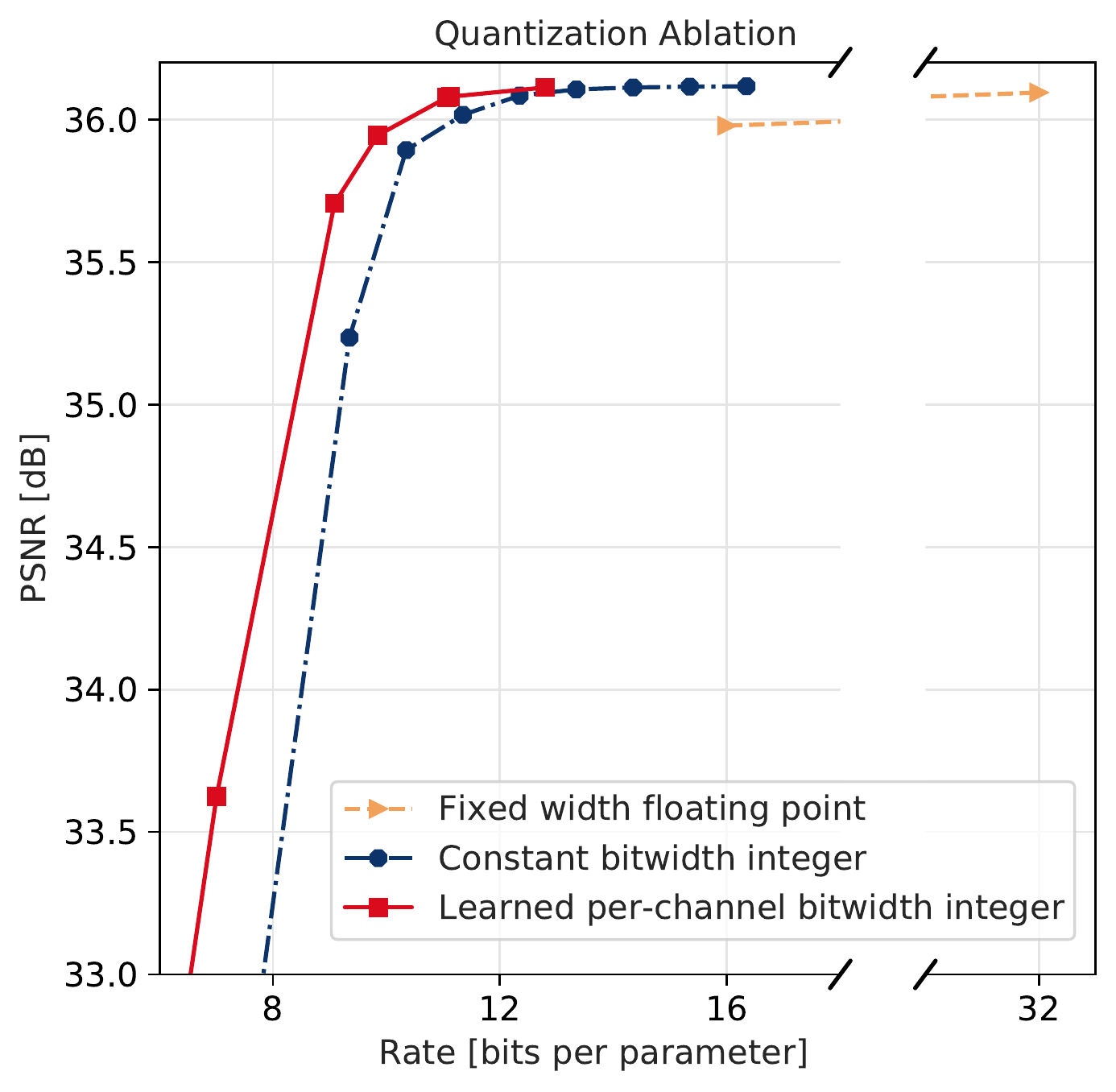}%
    \caption{Quantization ablation. Rate-distortion performance when quantizing a single model with different quantization techniques, learned per-channel quantization is the most performant. }
    \label{fig:quantization_ablations}
\end{figure}

In Fig.~\ref{fig:quantization_ablations} we show the effect of different quantization strategies on rate-distortion performance. 
Floating point quantization, the most naive quantization baseline, typically only supports to 16 bits/parameter and is outperformed by the other two quantization methods which can give similar distortion at 10--11 bits/parameter. For all bitrates, learned per-channel quantization outperforms the fixed bitwidth quantization. The learned-bitwidth quantization which we use in our main models can quantize up to 8--9 bits/parameter on average with only a neglegible drop in distortion performance.
This quantization strategy allows the model to allocate a different bitwidth for each channel in every parameter, as required for the best rate-distortion tradeoff. This effect is illustrated in Fig.~\ref{fig:quantization_bitw}. This plot shows the distribution of learned bitwidths for each parameter in the network. Especially the first and the last layers of the network require quantization to higher bitwidths, while the other layers are quantized to around 8 bits/parameter.

\paragraph{Video compression}
%
\begin{figure}[t]
    \centering%
    \includegraphics[width=0.42 \textwidth]{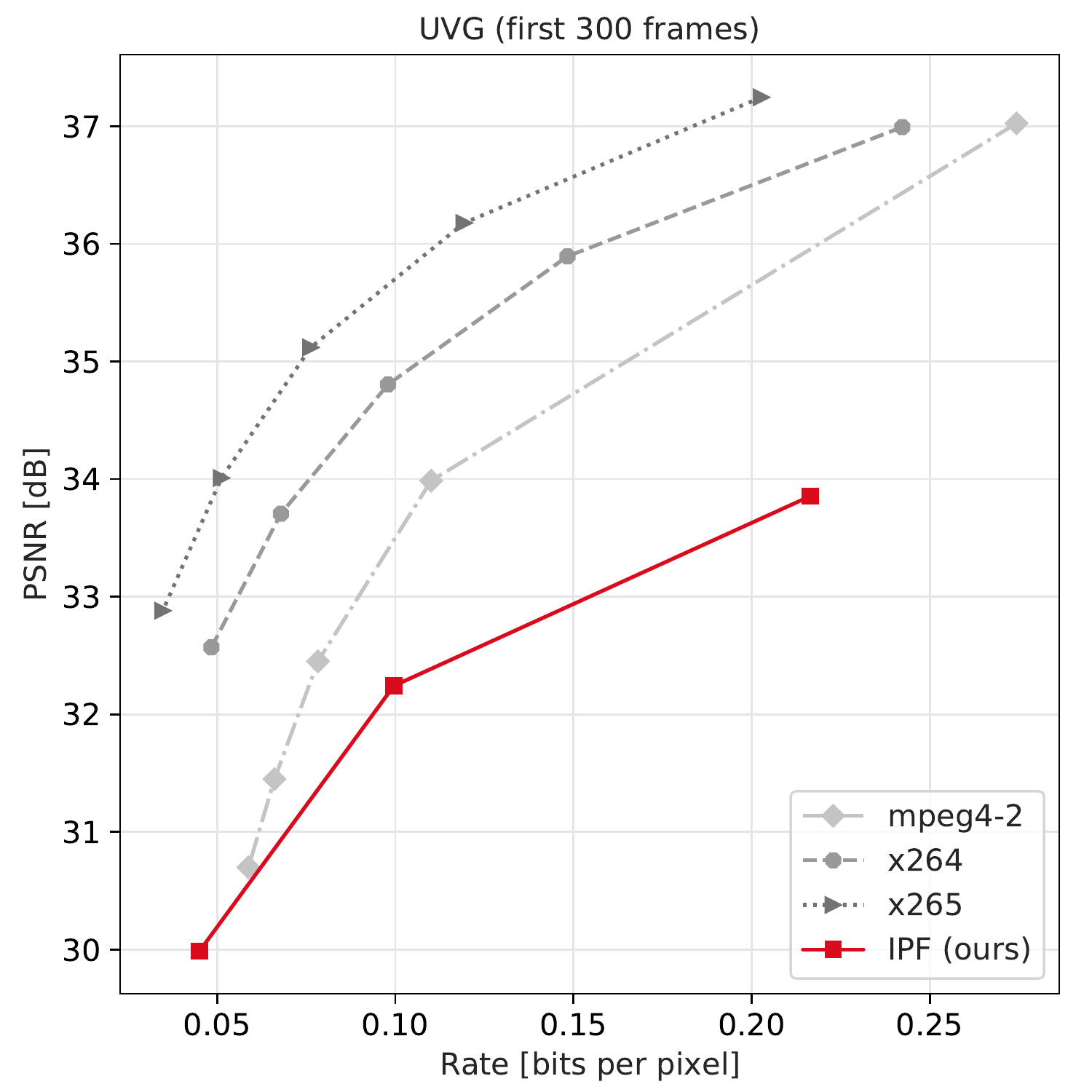}%
    \caption{Overall video compression performance of IPF and three types of baselines on the UVG dataset. Due to computational constraints we only show result of the first 300 frames. }
    \label{fig:rd}
\end{figure}

We compress the 7 videos from the UVG-1k dataset~\cite{uvg}, which have a Full-HD resolution ($1920 \times 1080$ pixels). As the content and style of these sequences do not change over the video, we only use the first half of each sequence (i.\,e.\ the first 300 frames) to save compute resources. We use three different architectures, corresponding to different working points on the rate-distortion curve. We specify the I-frame codec sizes to roughly cover the range of bit-rates we are interested in. For each model, the flow and residual models are then specified to be $1/40$ the size of the I-frame codec, a ratio we empirically determined to give good rate-distortion performance. We describe further details of the setup in the supplementary materials. 

As baselines we compare to the popular classical codecs mpeg4-2~\cite{mpeg4-2}, H.264~\cite{x264}, and H.265~\cite{x265} in the ffmpeg implementation~\cite{ffmpeg, libx264, libx265}. For our method and all baselines we use a GoP size of 5 frames and operate in the low-delay setting with only I-frames and P-frames. We trained our pipelines on TeslaV100 GPUs for approximately 300 GPU hours for each video.We show the average performance over all videos in Fig.~\ref{fig:rd}. Our method is able to compete with mpeg4-2 at low bit rates, but it is still clearly behind H.264 and H.265. 

\begin{figure}[t]
    \centering%
    \includegraphics[width=0.42\textwidth]{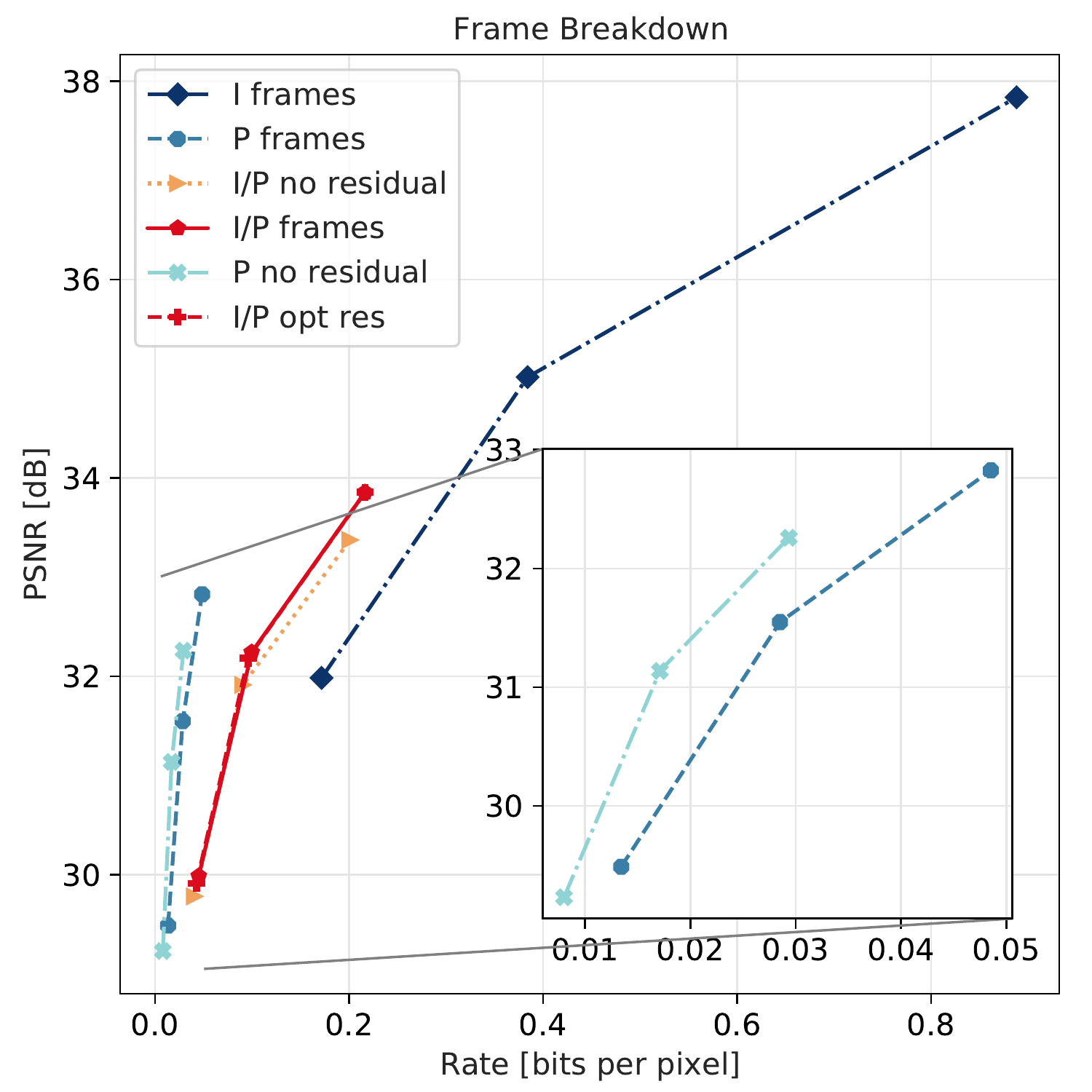}%
    \caption{Breakdown of performance for various frame types. The overall performance (Fig.~\ref{fig:rd}) is shown in solid red. }
    \label{fig:frame_bd}
\end{figure}

To gain some insights into this result, we breakdown the rate-distortion performance per type of frame in Fig.~\ref{fig:frame_bd}. The red solid line indicates our average performance as in Fig.~\ref{fig:rd}. This is averaged over I-frames (dot dashed dark blue) and P frames (dashed light blue). The I-frame are 2 - 5 dB higher in PSNR than the P-frames, depending on the model size, while being 20 times as expensive to compress (considering the combined rate of flow and residual, each of which is 1/40 size of the I-frame). 

The lower PSNR for P-frames can be partially attributed our use of much smaller models to model the P-frames. Furthermore, as discussed in Sec.~\ref{sec:implicit_video_representations}, our flow model is tasked with reproducing the P-frame single handedly. Thus it becomes a question whether the residual model is always needed. Thus we show the alternate scenario where we do not transmit the residual model both for overall (yellow), and P-frame only (cyan). In the inset we can see that the residual model roughly doubles the rate of the P-frames with a slight improvement in PSNR. While for the P-frames this means a sub-par distortion result, the overall performance with residual (red) still improves upon without (yellow). 

Finally, to take the best possible result regarding flow and residual, we dynamically decide whether to include the residual model in the bitstream as follows. For each GoP of each video, we compute the $RD$ loss with and without the residual model, and decide whether to include residual model for the given GoP based on the comparison. Although this breaks the low-delay setting for encoding, it still allows low-delay decoding. The decision of whether to include the residual model adds only a single bit per GoP to the bit stream. The result of this dynamic approach is shown in dot dashed red line, which we see slightly outperforms codecs with and without residual models. 

\section{Conclusion}

While neural compression algorithms are beginning to outperform classical codecs, they suffer from severe practical disadvantages. Not only do they require training datasets that match the data expected at test time, but they also require large pretrained neural networks on the receiver side. While such networks greatly aids rate-distortion performance, it presents an obstacle to practical use, especially on device.

We propose an image and video compression method based on implicit neural representations that avoids these obstacles. Each image is compressed by training a neural networks to learn its pixel content, quantizing the network weights, and entropy-coding them to the bitstream. For videos we introduced a frame-by-frame scheme that leverages the continuous nature of implicit representations to perform motion compensation without requiring hardware-unfriendly operations like interpolation. 

While our method is not yet competitive with the rate-distortion performance of state-of-the-art codecs, we demonstrate that it can nevertheless compress image and video data efficiently. Given its practical advantages, we believe it can serve as a step towards self-contained learning-based codecs that are deployable in real life use cases. 


\bibliographystyle{ieee_fullname}
\bibliography{arxiv}

\appendix{}

\section{Video demonstration}
\label{sec:video}

We include reconstructions of the ``Bosphorus'' sequence from the UVG dataset~\cite{uvg} for all three IPF models in the supplementary files. The videos are 300 frames at 120fps.

\section{Detailed results}

\subsection{Per-video breakdown}
\label{sec:per_video}

\begin{figure*}[t]
    \centering%
    \includegraphics[width=0.85\textwidth]{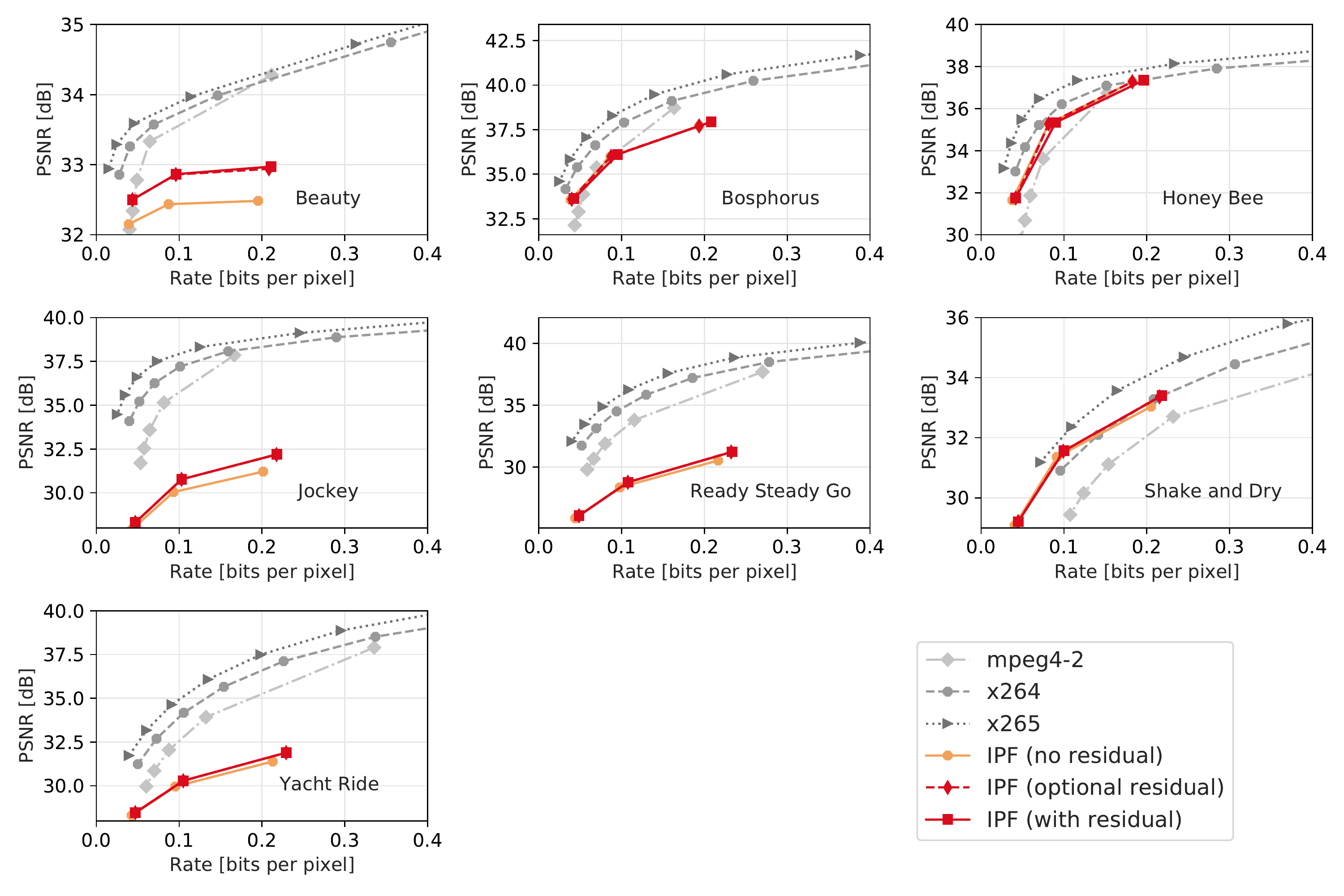}%
    \vskip-10pt
    \caption{Performance of our video codec on the different video sequences in the UVG dataset~\cite{uvg}. We show two versions of our codec: the default version with optional residual networks in red and a flow-only version without residual networks in yellow. }
    \vskip-8pt
    \label{fig:pervideo}
\end{figure*}

In Fig.~\ref{fig:pervideo} we show breakdown of our method against baselines for each video in the UVG dataset. Similar to overall result in the main text, we show the full model with residuals, no residuals, and a model with optional residual components. We see clear difference between much static videos such as Honey Bee and more dynamic videos such as Jockey. Our method performs better compared to the baselines on the static videos than on dynamic videos. Residual component leads to more improvement on dynamic videos. 

\subsection{Quantization details}

\begin{figure}[b]
    \centering%
    \includegraphics[width=0.46 \textwidth]{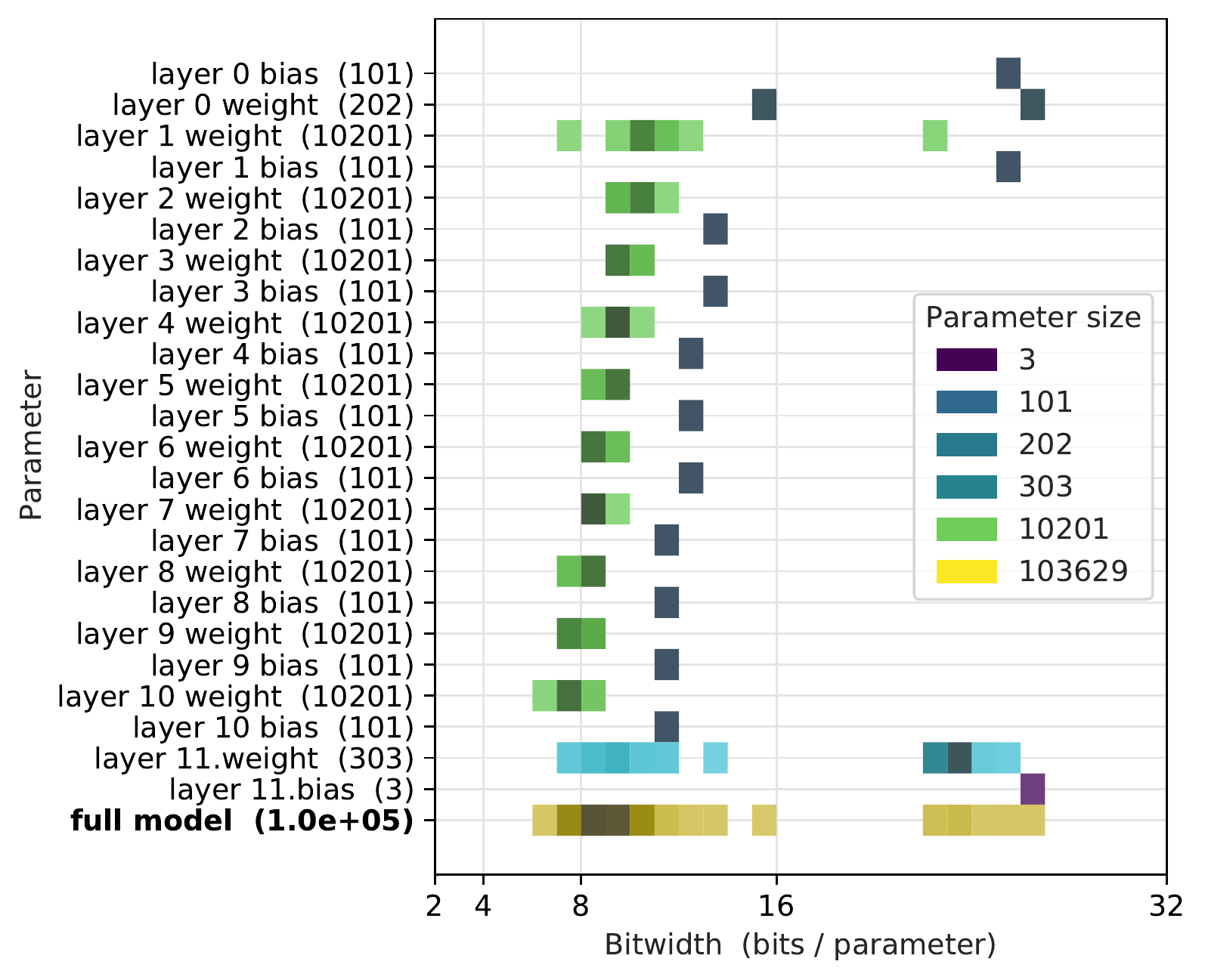}%
    \caption{The learned bitwidths for the implicit representation. Number of parameters in each layer are indicated in brackets. }
    \label{fig:quantization_bitw}
\end{figure}

Here we present the learned bitwidths for the implicit model.  Figure \ref{fig:quantization_bitw} shows the learned bitwidths for a single model (the model used to compress the image in Fig.~\ref{fig:perceptual}). It can be seen that the model learns an efficient bit-width of 8.1 bits/parameter. Furthermore, we can see that the model dynamically learns to allocate bit-width, spending more bits on small parameters such as biases, or on ``important'' layers such as the first and the last layer.

\subsection{Time implicit SIREN (SIREN3D)}
\label{sec:SIREN3D}

\begin{figure}[b]
    \centering
    \includegraphics[width=0.42\textwidth]{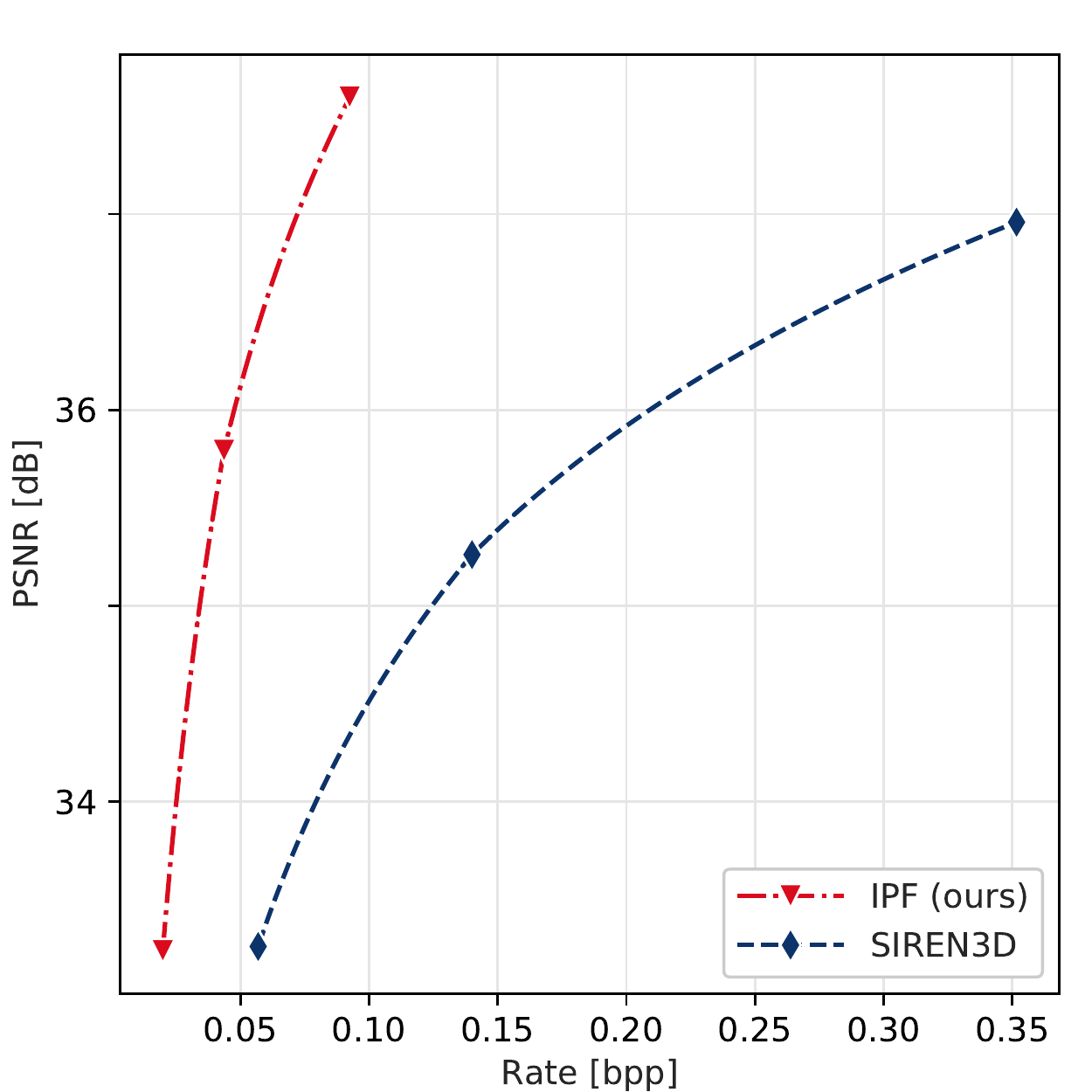}%
    \caption{Compression performance comparison on the first 25 frames of Bosphorus video from the UVG-1k dataset.}
    \label{fig:arch_ablation}
\end{figure}

When generalizing implicit neural compression models from two-dimensional images to video data, a natural choice is to incorporate the time axis as a new implicit dimension. The resulting architecture, which we here refer to as SIREN3D, operates directly on an entire group of pictures.

In Fig.~\ref{fig:arch_ablation} we compare the average performance of this model on the first 5 GoPs, each consisting of 5 frames, to that of our IPF model. SIREN3D underperforms IPF in all rate ranges tested. 

The careful reader may notice the rate differences between models. The initial SIREN3D models are roughly the same size as that of the IPF I-frame models (see Sec.~\ref{sec:hyperparams} for details). However, our quantization procedure uses the same value of $\beta$ in its optimization objective~ (Eq.~\eqref{eq:implicit_RD_loss}), thus the resulting rates are somewhat larger than that of their IPF counterparts. 

\section{Implementation details}
\label{sec:hyperparams}
\begin{table*}[b]
    \centering
    \small
    \begin{tabular}{lrrrrr}
        \toprule
        Dataset & Working point & Layers & Channels & Parameters & Parameters per pixel \\
        \midrule
        CLIC && 12 & 101 & 103\,629 & 0.029 \\
        \midrule
        Kodak & 1 & 5 & 20 & 1\,383 & 0.004 \\
         & 2 & 5 & 30 & 2\,973 & 0.008 \\
         & 3 & 10 & 28 & 6\,667 & 0.017 \\
         & 4 & 10 & 40 & 13\,363 & 0.034 \\
         & 5 & 13 & 49 & 27\,247 & 0.069 \\
         & 6 & 13 & 59 & 39\,297 & 0.100 \\
         & 7 & 13 & 66 & 49\,041 & 0.125 \\
    \bottomrule
    \end{tabular}
    \caption{Architecture details of SIREN models for image compression experiments.}
    \label{tab:arch_images}
\end{table*}

\begin{table*}[b]
    \centering
    \small
    \begin{tabular}{llll}
        \toprule
        Hyperparam & SIREN & uSIREN & SIREN3D\\
        \midrule
        Layers & 12 & 13 & 13\\
        Architecture & SSS-SSS-SSS-SSC & SSS-SSS-SSS-SUS-C & SSS-SSS-SSS-SSS-C\\
        \midrule
        {\it IPF--small} & & &\\
        Channels & 50 & 47 & 45\\
        Parameters & 25803 & 25101 & 25158\\
        \midrule
        {\it IPF--medium} & & &\\
        Channels & 79 & 76 & 72\\
        Parameters & 63677 & 64831 & 63579\\
        \midrule
        {\it IPF--large} & & & \\
        Channels & 127 & 121 & 116\\
        Parameters & 163325 & 163111 & 163679\\
    \bottomrule
    \end{tabular}
    \caption{Architecture details of SIREN, uSIREN and SIREN3D models for the video compression experiments.   For the architectures, we use S to denote siren layer (linear layer with sinusoidal activation), U to denote upsampling, while C stands for a normal $1 \times 1$ convolutional layer with relu activations. The hyphens are for readability only and does not have architectural meanings.}
    \label{tab:arch}
\end{table*}

\begin{table*}[tbp]
    \centering
    \small
    \begin{tabular}{lrrr}
        \toprule
        Hyperparameter & Steps (thousands) & Initial learning rate & Final learning rate\\
        \midrule
        Initial I-frame & 180 & $1\cdot 10^{-4}$ & $1\cdot 10^{-5}$ \\
        Other I-frame & 80 & $1\cdot 10^{-4}$ & $1\cdot 10^{-5}$ \\
        Initial flow training & 20 & $1\cdot 10^{-4}$ & $1\cdot 10^{-5}$ \\
        Initial flow quantization & 3 & $5\cdot 10^{-6}$ & $1\cdot 10^{-6}$ \\
        Other flow quantization & 20 & $1\cdot 10^{-4}$ & $1\cdot 10^{-6}$ \\
        Residual training & 20 & $1\cdot 10^{-4}$ & $1\cdot 10^{-5}$ \\
        Residual quantization & 3 & $2\cdot 10^{-5}$ & $1\cdot 10^{-7}$ \\
    \bottomrule
    \end{tabular}
    \caption{Learning rate scheduling in all stages of IPF training. All learnign rates are decayed exponentially from initial learning rate to final learning rate over given number of steps. }
    \label{tab:lr_schedule}
\end{table*}

\subsection{Image compression}

\paragraph{Architecture}

The architectures of the implicit models used in our image compression experiments are summarized in Tbl.~\ref{tab:arch_images}. Our choice for the Kodak dataset is based on the hyperparameters described in Ref.~\cite{dupont2021coin}, except for two differences. Firstly, on the very last layer we use ReLU activations while \cite{dupont2021coin} uses identity/no activations. Secondly, Ref.~\cite{dupont2021coin} use networks with one more layer than they describe in their appendix; we follow their description in the paper, not their implementation, and thus use one layer less in every model. We find that our models are comparable in distortion performance to theirs even with one fewer layer. In addition, we add two larger models aimed at a higher-quality setting.
 
\paragraph{Training hyperparameters}

We first train the implicit models with Adam for 100\,000 steps with a learning rate decaying exponentially from $1 \times 10^{-4}$ to $5 \times 10^{-6}$. Then we train the quantized models for an additional  $25\,000$ steps, using a constant learning rate of $2 \times 10^{-5}$ and weighting the bits per parameters in the loss function with $\beta = 10^{-4}$ for the lower-bitrate models and $\beta = 3 \times 10^{-5}$ for the higher-bitrate models.

\subsection{Video compression}

\paragraph{Architecture}
The architecture hyper-parameters of our I-frame SIREN, and uSIREN~(see Sec.~3.2 in the main text), as well as an alternative video model SIREN3D~(see Sec.~\ref{sec:SIREN3D}) are shown in Tbl.~\ref{tab:arch}. For a fair comparison, the architectures are chosen with a similar number of learnable layers. The number of channels are then chosen so that the total number of parameters match. Three model sizes (small, medium, and large) are explored. For simplicity we keep the number of layers the same across this size range. Not shown are the architectures of flow models. For the medium and large IPF, we use as flow model a 6 layer siren with 32 channels. For the small IPF models, we use 6 layers with 24 channels.

\paragraph{Training hyperparameters}

In this section we detail some of our training setup for the full IPF. For all stages we optimize towards a rate-distortion objective based on the MSE as distortion metric and with $\beta=10^{-4}$. We use the Adam optimizer throughout. Table~\ref{tab:lr_schedule} details the learning rate schedules.

We make our training more efficient by initializing each I-frame model except for the very first at the implicit model representing the previous I-frame. With this initialization, we only need to train for 80k steps.

Similarly, all flow models (except for the first in each GoP) are initialized to the flow model from the previous P-frame. We then skip the unquantized pretraining phase and directly optimize with quantization.

\section{Baselines}
\label{app:baselines}

\paragraph{CLIC}
In our first image compression analysis on an image from the CLIC 2020 Challenge (see Fig.~\ref{fig:perceptual}, we compare to a JPEG baseline generated with Pillow~\cite{pillow} with subsampling disabled.

\paragraph{Kodak}
On the Kodak image dataset, we compare to COIN, JPEG, JPEG2000, and BPG baselines as reported by Ref.~\cite{dupont2021coin}. 

\paragraph{Video}
We compare to three popular classical codecs (mpeg4-part2, H.264, H.265) in their ffmpeg (x265, x266) implementations~\cite{ffmpeg, libx265}.
All methods are restricted to only I-frames and P-frames and a fixed GoP size of 5. We use the ffmpeg ``medium'' encoder preset.

\end{document}